\documentclass{article}
\usepackage{color}
\usepackage{spconf,amsmath,graphicx,amssymb,subfigure}
\usepackage{cite}

\title{Semantically proportional patchmix for few-shot learning}
\name{Jingquan Wang$^{\dagger}$, Jing Xu$^{\dagger}$, Yu Pan$^{\dagger}$, Zenglin Xu$^{{\dagger},{\ddagger},*}$\thanks{* means the corresponding author.
}
\thanks{This work was partially supported by the National Key Research and Development Program of China (No.2018AAA0100204), and a key program of fundamental research from Shenzhen Science and Technology Innovation Commission(No. JCYJ20200109113403826).
}
}
\address{$^{\dagger}$Harbin Institute of Technology (Shenzhen), Shenzhen, Guangdong, China\\
$^{\ddagger}$PengCheng Laboratory, Shenzhen, Guangdong, China}
\begin{document}
%\ninept
%
\maketitle
\begin{abstract}
Few-shot learning aims to classify unseen classes with only a limited number of labeled data. Recent works have demonstrated that training models with a simple transfer learning strategy can achieve competitive results in few-shot classification. Although excelling at distinguishing training data, these models are not well generalized to unseen data, probably due to insufficient feature representations on evaluation. To tackle this issue, we propose Semantically Proportional Patchmix (\textbf{SePPMix}), in which patches are cut and pasted among training images and the ground truth labels are mixed proportionally to the semantic information of the patches. In this way, we can improve the generalization ability of the model by regional dropout effect without introducing severe label noise.
%Data mixing methods have proved valid in training a more general and transferable model for image classification recently. Inspired by that, we propose semantically proportional patchmix which can significantly boost the performance in practice. 
%increase the power of learned feature representations.
%Specifically, we swap patches of images randomly, forcing the network to attend on less discriminative parts of inputs, which lead to improved generalization in testing when only a few labeled samples are given. 
%The ground truth labels are mixed according to the semantic composition rather than the area of patches, we use class activation map (CAM) to estimate the patches' semantic to lessen the label noise after mixing.
% Furthermore, to lessen the label noise after mixing, we use class activation map (CAM) to estimate the semantic information of patches and get the ground truth mixed labels according to the semantic composition rather than simply the area of patches.
To learn more robust representations of data, we further take rotate transformation on the mixed images and predict rotations as a rule-based regularizer. 
%In addition, considering the scarcity of labeled samples, we take rotate transformation on the mixed images and use a rotation prediction task as rule-based regularizer to help learn generalizable features. 
Extensive experiments on prevalent few-shot benchmarks have shown the effectiveness of our proposed method.
\end{abstract}
\section{Introduction}
\label{sec:intro}

Deep learning has achieved significant progress in computer vision tasks. This success can be partly attributed to the availability of large-scale data~\cite{DBLP:journals/ijcv/RussakovskyDSKS15}. However, acquiring enough labeled data is infeasible in some situations due to event scarcity or expensive labor costs. Inspired by the human visual system that can learn new classes with only a few samples, 
%(e.g., a baby can recognize a panda with several glances),
few-shot learning (FSL) emerges as a promising method that aims to equip a learner with fast adaptability to novel concepts. For FSL, there are usually a base set and a novel set that share no class overlap. Then a model is trained on the base set with sufficient labeled data, and is further adapted to categorize the unseen classes in the novel set with scarce labeled data. In particular, the performance of the few-shot classification highly relies on the model's generalization ability.

Recent works\cite{DBLP:conf/iclr/ChenLKWH19,DBLP:conf/eccv/TianWKTI20} have shown that a model trained on the base set with standard supervision can achieve impressive results on the novel set. However, we argue that the learned model tends to be overly discriminative to the training data in the base set, leading to sub-optimal performance when evaluated on unseen classes. This is because that deep networks are likely to memorize some specific training statistics, resulting in overconfident prediction and poor generalization ability\cite{DBLP:conf/nips/ThulasidasanCBB19}. To solve this issue and make the model produce sufficient discriminative representations of unseen test data, it is essential to introduce uncertainty to the input data and to regularise the model in the training process.

Data mixing methods introduce uncertainty and reduce the risk of overly memorizing input data by blending original images and labels, which have proved valid in training a more general and transferable model for image classification recently\cite{DBLP:conf/wacv/Mangla0SKBK20,DBLP:conf/iccv/YunHCOYC19,DBLP:conf/aaai/HuangWT21}. In this paper, we propose a simple yet effective regularization method called Semantically Proportional Patchmix (\textbf{SePPMix}) to improve the performance of FSL. In SePPMix, we divide the input images into N$\times$N patches. The patches
are randomly cut from one image and pasted onto another image at their respective original locations, which prevents the model from learning over specific structures of training data. Furthermore, rather than directly using the area of patches, the label of the mixed image is generated by the semantic proportion of patches which is estimated by the class activation maps (CAMs)\cite{DBLP:conf/cvpr/ZhouKLOT16}. To increase the number of mixed samples and learn more robust representations of data, we rotate the mixed images and predict the rotated angle of the images as an auxiliary task. We conduct extensive experiments on two widely-used datasets, %showing that our SePPMix is better than other mix methods.
the empirical results show that the proposed method can significantly improve few-shot image classification performance over corresponding baseline methods. 
Remarkably, under the same experiment setting,
the proposed method improves 1-shot and 5-shot tasks by nearly \textbf{6\%}
and \textbf{5\%} on MiniImageNet and CUB, respectively.
%Considering the fewer training samples in training stage, directly determine the ground-truth label according to the proportion to the number of pixels of combined images may introduce noise. Thus we define our ground-truth labels via the semantic composition, concretely, we exploit class activation map(CAM) to define the patches' proportion. To further increase the variety and quantity of training data, we rotate the mixed images like\cite{DBLP:conf/iclr/GidarisSK18}  and predict the rotate angle of the images. Extensive expriments on two benchmarks show the effectiveness of our proposed method.
% In this paper, we propose a simple but effective mix method to improve the performance of FSL. Different from CutMix, we divided the input images into N$\times$N patches, the patches are randomly swapped between two images, our mix method allow the quantity of patches not to be unique and patches' location to be more extensive than CutMix. Considering the fewer training samples in training stage, directly determine the ground-truth label according to the proportion to the number of pixels of combined images may introduce noise. Thus we define our ground-truth labels via the semantic composition, concretely, we exploit class activation map(CAM) to define the patches' proportion. To further increase the variety and quantity of training data, we rotate the mixed images like\cite{DBLP:conf/iclr/GidarisSK18}  and predict the rotate angle of the images. Extensive expriments on two benchmarks show the effectiveness of our proposed method.

\section{RELATED WORKS}
\label{sec:format}

\textbf{Few-shot learning} aims to build
a model using available training data of base classes that
can classify unseen novel classes with few labeled examples in each class. One popular paradigm to tackle few-shot classification is meta-learning, which can be roughly divided into two streams, optimized-based methods~\cite{DBLP:conf/icml/FinnAL17,DBLP:journals/corr/abs-1803-02999,DBLP:conf/iclr/RusuRSVPOH19} and metric-based methods~\cite{DBLP:conf/nips/VinyalsBLKW16, DBLP:conf/nips/SnellSZ17, luo2021rectifying, DBLP:conf/cvpr/ZhangCLS20}. The first class of methods, i.e., learning to learn methods, aiming to learn a suitable model initialization, so that the model can achieve fast adaption with a few labeled examples.
%In the former class of methods, %MAML\cite{DBLP:conf/icml/FinnAL17} aims to finda good initialization of model parameters so that it can be quickly adapted to novel data with fewer steps, Reptile\cite{DBLP:journals/corr/abs-1803-02999} is simpler than MAML which removes re-initialization for each task. LEO\cite{DBLP:conf/iclr/RusuRSVPOH19} learns a lower dimension latent space to reduce the complexity. 
The second category is learning to compare methods, targeting at projecting input samples into a specific embedding space where data from different classes can be
distinguished by using distance metrics. Apart from these, recent works\cite{DBLP:conf/iclr/ChenLKWH19,DBLP:conf/eccv/TianWKTI20} adopt the transfer learning strategy to train the model on all the base classes, then use it to extract the feature representation of input data on the novel classes, showing competitive performance. Our work follows this school of thought and further improves the generalization of the model.
%Recent works\cite{DBLP:conf/iclr/ChenLKWH19,DBLP:conf/eccv/TianWKTI20,DBLP:conf/iclr/DhillonCRS20} use standard transfer-learning strategy show surprisingly competitive performance, that first pre-train the embedding on the whole base set, then finetune on the novel set.  

\textbf{Data augmentation} has been widely used to train deep learning models. 
%Directly applying simple and inexpensive transformations (e.g., rotation, scaling) increase the amount and diversity of exiting data. 
Regional-erasing methods~\cite{DBLP:conf/iccv/SinghL17,DBLP:journals/corr/abs-1708-04552} erasing part of the training images, aiming to encourage the networks better utilize the entire context of the data.
%Another line is data mixing methods:
Mixup\cite{DBLP:conf/iclr/ZhangCDL18}  generates images by linearly combining data
and fusing their labels using the same coefficients and shows its generalization ability in large datasets; Manifold Mixup\cite{DBLP:conf/icml/VermaLBNMLB19} applies Mixup to the intermediate layer, leading to smoother decision boundaries which benefit feature representation; Cutmix\cite{DBLP:conf/iccv/YunHCOYC19} combines the advantages of Cutout\cite{DBLP:journals/corr/abs-1708-04552} and Mixup, which produces a new image by cutting out one random image patch and pasting it to another image. Different from Cutmix, Snapmix\cite{DBLP:conf/aaai/HuangWT21} generates the target label for the mixed image by estimating its intrinsic semantic composition, which achieves top-level performance in fine-grained data. However, these data mixing methods are not specifically designed for limited data, which may result in suboptimal performance in FSL tasks.
\section{METHOD}
\label{sec:pagestyle}
\subsection{Problem Formulation}
Few-shot learning usually consists of two disjoint datasets $\mathcal D_b$ and $\mathcal D_n$.
%$\mathcal D_b$ = 
%$\{x_b^i, y_b^i\}_{i=1}^{C_b}$ and $\mathcal D_{n}$ = $\{x_n^j, y_n^j\}_{j=1}^{C_n}$, where $C_b$ and $C_n$ are the number of base classes and novel classes respectively. 
A model is firstly trained on the sufficient labeled dataset $\mathcal D_b$, and then performs the FSL task based on the novel dataset $\mathcal D_n$. The common way to build the FSL task is called $N$-way $K$-shot that should classify $N$ classes sampled from $\mathcal D_n$ with $K$ labeled data in each class. The few labeled data is called support set $\mathcal S = \{x_n^j, y_n^j\}_{j=1}^{N \times K}$ that contains N classes with each class K examples. The performance is evaluated on the query set $\mathcal Q = \{x_n^j, y_n^j\}_{j=N \times K + 1}^{N \times K + H}$, where the $H$ denotes the number of unlabeled examples and the data in $\mathcal Q$ is sampled from the same $N$ classes in each episode. Most meta-learning methods\cite{DBLP:conf/nips/VinyalsBLKW16,DBLP:conf/icml/FinnAL17,DBLP:conf/nips/SnellSZ17} adopt the same episode training scheme as evaluation when trained on $\mathcal D_b$, making training and evaluation stage in the same situation. However, recent works\cite{DBLP:conf/iclr/ChenLKWH19,DBLP:conf/iclr/DhillonCRS20,DBLP:conf/eccv/TianWKTI20} show that training the model on the whole $\mathcal D_b$ in a supervised manner is efficient for the FSL task. We follow the same idea and utilize our proposed method to train a more generalized embedding network on $\mathcal D_b$, which performs better than baseline when tested on $\mathcal D_n$.

\begin{figure}[t]
 
\centering
\includegraphics[scale=0.46]{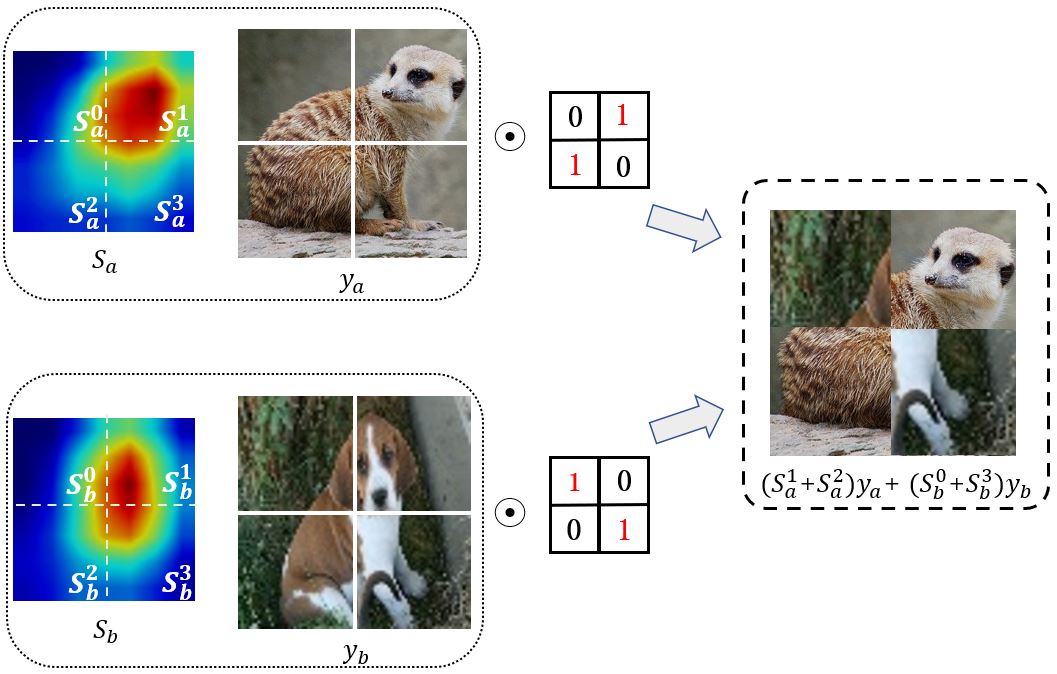}  %这个是图片的绝对路径
\caption{An overview of SePPMix. $S_a$ and $S_b$ are the semantic information maps of input image $x_a$ and $x_b$ respectively.}
\label{fig3}
\end{figure}

\subsection{Framework Overview}
In this work, we consider the simple but effective procedure as \cite{DBLP:conf/eccv/TianWKTI20}, during the first stage, we train the embedding network $f_{\phi}$ on $\mathcal D_b$,
\begin{equation}
{\phi} = \mathop{\arg\min}_{\phi}{\mathcal L_{base}(\mathcal D_b;\phi)},
\end{equation}
where {$\mathcal L_{base}$} is the loss function, $\phi$ is the parameters of the embedding model. Then the pre-trained $f_{\phi}$ is fixed and used as the feature extractor in $D_n$. A linear classifier $g_{\theta}$ is firstly  trained on the extracted features of $\mathcal S$,
\begin{equation}
{\theta} = \mathop{\arg\min}_{\theta}{\mathcal L_{base}(\mathcal S;\theta,\phi)}+R(\theta),  
\end{equation}
where $\theta$ contains weight and bias terms and $R$ is the regulation term. Then $g_{\theta}$ can be leveraged as predictor on the features of $\mathcal Q$. %Importantly, we propose a new mix method to learn more general representation for FSL tasks. Additionally, we  rotate the mixed samples to increase diversity and quantity and predict the degree of rotation as an auxiliary loss, which results in more transferable embeddings. 
\subsection{Sample Generation via SePPMix}
%In few-shot learning, the category difference usually resides in parts of the object due to the scarce of training data, making generalization ability play an important role. 
In few-shot learning, 
%the category difference usually resides in base and novel classes which makes FSL aims to learn the model on seen classes capable of generalization ability to unseen classes.
SePPMix can better improve the model generalization than other data mixing methods.
%Therefore, we hypothesize that using cut-and-paste mechanism is more favorable in augmenting few-shot data. 
%Therefore, we proposed SePPMix to augment base data using cut-and-paste operations, which can improve the generalization ability of the learned model. 
On the one hand, it increases the uncertainty and diversity of the training data; on the other hand, it generates more accurate labels of mixed data without introducing severe noise.
%Our SppMix divides image into N $\times$ N patches. Every patch of the image is randomly replaced by the patch corresponding to the position of another image. Motivated by the work\cite{DBLP:conf/aaai/HuangWT21} that used class activation maps(CAMs) to estimate the semantic composition of a mixed image in fine-grained data, we exploit CAM to estimate the semantic proportion of patches as the guidance to fuse new labels.
Fig.~\ref{fig3} illustrates the procedure of our proposed method.

Formally, let $x\in{\mathbb R^{3 \times h \times w}}$ denotes the training image and $y$ is its corresponding label. SePPMix create a new training sample $(\hat{x},\hat{y})$ given two distinct samples ($x_a$,$y_a$) and ($x_b$,$y_b$). The generation of mixed image is defined as
\begin{equation}\label{eq3}
\hat{x} = M \odot x_a + (\mathbf 1-M) \odot x_b,\\
\end{equation}
where $M \in\{ 0,1\}^{N\times N}$ denotes a binary mask indicating which patches belong to either of the two images. The $\mathbf 1$ is a binary mask filled with ones and the probability of its appearing is 0.5, and $\odot$ denotes the element-wise multiplication. 

When using the original data of the base set for training, the different patches in the image contain varying levels of information that are valid for the classification task. 
%Some patches are more informative or even critical, while others are not. 
To generate a more accurate label,  we calculate the class activation maps of images as the indicators of semantic
information and use them to estimate how a patch correlates with its label, 
%we need to estimate the semantic composition of the mixed image. We use the attention method \cite{DBLP:conf/cvpr/ZhouKLOT16} to compute the class activation maps of input images,
which has been proved useful to measure the semantic relatedness of each original image patch to the corresponding label \cite{DBLP:conf/aaai/HuangWT21}.
%Motivated by\cite{}, we estimate the semantic composition of the new sample and compute class activation map using attention method\cite{}, which has been demonstrated useful to interpret how a region correlates with a semantic class. 
% We use $F(x) \in \mathbb R^{c\times h\times w}$ to denote the output feature of the last convolutional layer of image x, then we can obtain its 
Given an image $x$, the class activation map can be calculated by
\begin{equation}
CAM(x) = {\Phi}(\sum_{l=1}^cw_y^lF_l(x)),
\end{equation}
where $\Phi(\cdot)$ is the operation that upsamples a feature map to match the dimensions of the input image. $F(\cdot)$ denotes the feature extractor and $F_l(x)$ denotes the $l^{th}$ feature map with input $x$, c is the number of feature maps. The $w_y$ is the classifier weight of class $y$, and for simplicity, the bias term is ignored. We obtain the semantic information map $S(x)$ by normalizing the $CAM(x)$ to sum to one. The semantic information map of $x$, $S(x)$, is defined as
\begin{equation}
S(x) = \frac{CAM(x)}{sum(CAM(x))}.
\end{equation}

Finally, we can calculate the corresponding semantic label of the image produced using Eq.~\ref{eq3},
%we divide the images and its corresponding semantic percent map into N $\times$ N sub-patches respectively. M$\in$ $\{ 0,1\}^{N\times N}$ denotes a binary mask indicating each patch in the new sample and map belong to which of the two examples.
% M$\in$ $\{ 0,1\}^{N\times N}$ denotes a binary mask indicating that each patch in the mixed image belong to which of the two examples. 
%1 is a binary mask all filled with ones and the probability of it appearing is 0.5. The mixup image and corresponding label are defined as:
\begin{align}
&\rho_a = sum(M \odot S(x_a)), \\
&\rho_b = sum((\mathbf 1-M) \odot S(x_b)),\\
&\hat{y} = \rho_a y_a+\rho_b y_b,
\end{align}
where $\rho_a$ and $\rho_b$ are the weights corresponding to label $y_a$ and $y_b$ respectively. The newly generated image label in Fig ~\ref{fig3} depicts this process. In this way, not only can the model learn rich visual features, but also prevent introducing heavy noise in the augmented data, especially when in the few-shot situation.

\subsection{  Training SePPMix}
Our SePPMix is applied in the training stage to learn better representations with generalization ability. Given a new sample $(\hat{x},\hat{y})$ generated by SePPMix, we rotate it and predict the angles as an auxiliary task to learn more generalizable features. $\hat{x}^r$ denotes the mixed image rotated by r degrees, r $\in C_R$=$\{{0}^{\circ},{90}^{\circ},{180}^{\circ},{270}^{\circ}\}$. The loss $\mathcal L_m$ for training the image classification task using mixed images and the auxiliary loss $\mathcal L_r$ for predicting the rotation angle are defined as
\begin{align}
&\mathcal L_m = \sum_{x \in D_b}\sum_{r \in C_R}\mathcal L_{ce}[f_\phi(\hat{x}^r),\hat{y}],\\
&\mathcal L_r = \frac{1}{|C_R|}\sum_{x \in D_b}\sum_{r \in C_R}L_{ce}[g_r(f_\phi(\hat{x}^r)),r],
\end{align}
where $\mathcal L_{ce}$ is standard cross-entropy loss function and $g_r(\cdot)$ is a 4-way linear classifier. $|C_R|$ denotes the cardinality of $C_R$. Then the overall loss in training stage is
\begin{equation}
\mathcal L_{base} = \alpha \mathcal L_m + \beta \mathcal L_r.
\end{equation}
where $\alpha$ and $\beta$ are the weighting factors. For evaluation, the embedding network $f_\phi$ is fixed and tested on FSL tasks consists of $\mathcal S$ and $\mathcal Q$ sampled from $\mathcal D_n$ which plays a role as feature extractor, and we obtain embeddings of the images in $\mathcal S$ and $\mathcal Q$. We follow the implementation of\cite{DBLP:conf/eccv/TianWKTI20}, which trains a logistic regression classifier based on the embeddings of images in $\mathcal S$ and their corresponding labels, and uses the classifier to predict the labels of images in $\mathcal Q$.

\section{EXPERIMENTS}
\label{sec:typestyle}
\subsection{Datasets}
We perform our experiments on two widely used datasets in FSL, i.e., miniImageNet\cite{DBLP:conf/iclr/RaviL17} and CUB\cite{WelinderEtal2010}. The miniImageNet is a subset of ILSVRC-12\cite{DBLP:journals/ijcv/RussakovskyDSKS15}, including 100 distinct classes, each of which contains 600 images of size 84 $\times$ 84. We adopt the common setup introduced by\cite{DBLP:conf/iclr/RaviL17}, which split the categories into 64,16,20 classes for training, validation and evaluation respectively. The CUB dataset consisting of 11,788 images of size 84 $\times$ 84 from 200 bird classes. We follow the splits from~\cite{DBLP:conf/iclr/ChenLKWH19}, where 100 classes are used for training, 50 for
validation, and 50 for testing.
\begin{table}[t]
\centering
\small
\caption{\label{comp1}Average 5-way few-shot classification accuracies (\%) with 95{\%}
confidence intervals on miniImageNet dataset.}
\begin{tabular}{llll}
\hline
{\textbf{Method}} & \multicolumn{1}{c}{\textbf{Backbone}} & \multicolumn{1}{c}{\textbf{1-shot}} & \multicolumn{1}{c}{\textbf{5-shot}} \\ \hline
\rule{-2pt}{10pt}
MAML\cite{DBLP:conf/icml/FinnAL17}&32-32-32-32&48.70$\pm$1.84&63.11$\pm$0.92\\

MatchingNet\cite{DBLP:conf/nips/VinyalsBLKW16}&64-64-64-64&43.56$\pm$0.84&55.31$\pm$0.73\\
ProtoNet\cite{DBLP:conf/nips/SnellSZ17}&64-64-64-64&49.42$\pm$0.78&68.20$\pm$0.66\\
TADAM\cite{DBLP:conf/nips/OreshkinLL18}&ResNet-12&58.50$\pm$0.30&  76.70$\pm$0.30\\
MTL\cite{DBLP:conf/cvpr/SunLCS19}&ResNet-12&61.20$\pm$1.80& 75.50$\pm$0.80\\
MetaOptNet\cite{DBLP:conf/cvpr/LeeMRS19}&ResNet-12&62.64$\pm$0.61&78.63$\pm$0.64\\
Fine-tuning\cite{DBLP:conf/iclr/DhillonCRS20}&WRN-28-10&57.73$\pm$0.62&78.17$\pm$0.49\\
VLCL\cite{luo2021boosting}&WRN-28-10&61.75$\pm$0.43&76.32$\pm$0.49\\
S2M2\cite{DBLP:conf/wacv/Mangla0SKBK20}&WRN-28-10&64.93$\pm$0.18&83.18$\pm$0.11\\
FEAT\cite{DBLP:conf/cvpr/YeHZS20}&ResNet-12&66.78$\pm$0.20&82.05$\pm$0.14\\
DeepEMD\cite{DBLP:conf/cvpr/ZhangCLS20}&ResNet-12&65.91$\pm$0.82&82.41$\pm$0.56\\
RFS-distill\cite{DBLP:conf/eccv/TianWKTI20}&ResNet-12&64.82$\pm$0.60&82.14$\pm$0.43\\

\hline
\rule{-2pt}{9pt}
RFS-simple\cite{DBLP:conf/eccv/TianWKTI20}&ResNet-12&62.02$\pm$0.63&79.64$\pm$0.44\\
\textbf{Ours}&ResNet-12&\textbf{66.98$\pm$0.81}&\textbf{83.88$\pm$0.54} \\    \hline                  
\end{tabular}

\end{table}

\subsection{Implementation Details}
Following\cite{DBLP:conf/eccv/TianWKTI20}, we use a ResNet-12 network as our backbone, which consists of 4 residual blocks with Dropblock as a regularizer. To generate the initial CAMs of the images, we train the network from scratch in $D_b$ at the beginning. In all experiments, we use Stochastic Gradient Descent (SGD) with a learning rate of 0.05, a momentum of 0.9, and a weight decay of 5$e^{-4}$. The training epoch is 65 epochs and the learning rate is decreased with a factor of 0.1 at 30, 45 and 60 epoch. The batch size is 64 and 16 on miniImageNet and CUB respectively. 
We empirically set the number of patches N=2, the coefficients $\alpha$ = 1 and $\beta$ = 0.5. We randomly sample 600 episodes to report the accuracies on both datasets.

\begin{table}[t]
\centering
\small
\caption{\label{comp2}Average 5-way few-shot classification accuracies (\%) with 95{\%}
confidence intervals on CUB. RFS-simple$^*$ indicates the model re-implemented by ours on CUB dataset.}
\begin{tabular}{llll}
\hline
{\textbf{Method}} & \multicolumn{1}{c}{\textbf{Backbone}} & \multicolumn{1}{c}{\textbf{1-shot}} & \multicolumn{1}{c}{\textbf{5-shot}} \\ \hline
\rule{-2pt}{10pt}
Baseline\cite{DBLP:conf/iclr/ChenLKWH19}&ResNet-18&65.51$\pm$0.87&82.85$\pm$0.55\\
Baseline++\cite{DBLP:conf/iclr/ChenLKWH19}&ResNet-18&67.02$\pm$0.90&83.58$\pm$0.54\\
MAML\cite{DBLP:conf/icml/FinnAL17}&ResNet-18&68.42$\pm$1.07&83.47$\pm$0.62\\
MatchingNet\cite{DBLP:conf/nips/VinyalsBLKW16}&ResNet-12&71.87$\pm$0.85&85.08$\pm$0.57\\
ProtoNet\cite{DBLP:conf/nips/SnellSZ17}&ResNet-12&66.09$\pm$0.92&82.50$\pm$0.58\\
Negative-Cosine\cite{DBLP:conf/eccv/LiuCLL0LH20}&ResNet-18&72.66$\pm$0.85&89.40$\pm$0.43\\
S2M2\cite{DBLP:conf/wacv/Mangla0SKBK20}&ResNet-18&71.43$\pm$0.28&85.55$\pm$0.52\\
DeepEMD\cite{DBLP:conf/cvpr/ZhangCLS20}&ResNet-12&75.65$\pm$0.83&88.69$\pm$0.50\\
\hline 
\rule{-2pt}{9pt}
RFS-simple$^*$\cite{DBLP:conf/eccv/TianWKTI20}&ResNet-12&72.78$\pm$0.86&87.24$\pm$0.50\\
\textbf{Ours}&ResNet-12&\textbf{78.55$\pm$0.77}&\textbf{91.81$\pm$0.41} \\    \hline               
\end{tabular}

\end{table}
%Following the data augmentation in \cite{}, we use the random crop, color jittering, and random horizontal flip when training the embedding network.

\subsection{Results}
We present the results of SePPMix on two representative benchmarks of the few-shot learning task. As detailed in Table~\ref{comp1}, our method outperforms all previous approaches on miniImageNet. Specifically, our approach achieves 4.96\% and 4.24\% improvement over our baseline RFS-simple\cite{DBLP:conf/eccv/TianWKTI20} for 5-way 1-shot and 5-way 5-shot tasks respectively.
Table~\ref{comp2} illustrates the experimental results on CUB dataset. We re-implement the baseline RFS-simple\cite{DBLP:conf/eccv/TianWKTI20} on CUB. In the 5-way tasks, our method improves the accuracies by 5.77\% and 4.57\%  over the baseline on 1/5-shot respectively. Besides, our approach outperforms all the strong competitors and sets new records on 5-way 1/5-shot tasks on CUB.
\begin{table}[t]
\centering
\small
\caption{\label{comp3}Performance comparison between different mix methods and ablation study of our model, r indicates rotation.}
\begin{tabular}{l|ll}
\hline
\multicolumn{1}{c|}{} & \multicolumn{2}{c}{miniImageNet}\\
Method& \multicolumn{1}{c}{1-shot} & \multicolumn{1}{c}{5-shot}\\ \hline
baseline&\multicolumn{1}{c}{61.06} & \multicolumn{1}{c}{78.83}\\ \hline 
Mixup&59.12 (-1.94) &79.01 (+0.18)  \\ \hline
Cutmix&61.61 (+0.55) & 80.35 (+1.52) \\ \hline
Snapmix&62.52 (+1.46) &80.90 (+2.07)  \\
\hline
\hline
Patchmix&62.75 (+1.69) &80.69 (+1.86) \\ \hline
%rotation&63.26 (+2.20)&82.19 (+3.36)  \\ \hline
SePPMix (w/o r)&64.09 (\textbf{+3.03})&81.21 (\textbf{+2.38})  \\ \hline
SePPMix (w/ r)&66.98 (\textbf{+5.92}) &83.88 (\textbf{+5.05})  \\ \hline
\end{tabular}

\end{table}

\label{sec:majhead}
\begin{figure}[htbp]
\centering
\begin{minipage}[t]{0.19\linewidth}
\raggedright
\includegraphics[width=0.95\textwidth]{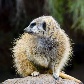}
\includegraphics[width=0.95\textwidth]{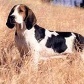}
\centerline{(a)}
%\caption{fig1}
\end{minipage}%
\begin{minipage}[t]{0.19\linewidth}
\raggedleft
\includegraphics[width=0.95\textwidth]{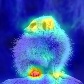}
\includegraphics[width=0.95\textwidth]{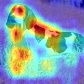}
\centerline{(b)}
%\caption{fig1}
\end{minipage}%
\begin{minipage}[t]{0.19\linewidth}
\raggedleft
\includegraphics[width=0.95\textwidth]{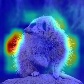}
\includegraphics[width=0.95\textwidth]{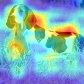}
\centerline{(c)}
%\caption{fig2}
\end{minipage}%
\begin{minipage}[t]{0.19\linewidth}
\raggedleft
\includegraphics[width=0.95\textwidth]{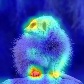}
\includegraphics[width=0.95\textwidth]{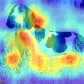}
\centerline{(d)}
%\caption{fig2}
\end{minipage}
\begin{minipage}[t]{0.19\linewidth}
\raggedright
\includegraphics[width=0.95\textwidth]{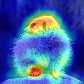}
\includegraphics[width=0.95\textwidth]{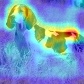}
\centerline{(e)}
%\caption{fig2}
\end{minipage}
\centering
\caption{\label{fig2}Class activation maps for different methods: (a) input, (b) baseline, (c) Mixup, (d) Cutmix, and (e) SePPMix. Here the warmer color indicates higher value.}
\end{figure}
\subsection{Ablations and Analysis}
We conduct an ablation study to compare SePPMix with other data mixing methods and assess the effects of different components of SePPMix. We report the results on both 5-way 1/5-shot tasks. In Table~\ref{comp3}, we train a simple baseline model without using data mixing augmentation. Notably, Mixup leads to worse performance in 5-way 1-shot task than the baseline, which indicates that linearly combining images and labels cannot improve the transfer and generalization abilities of the model in the FSL scenario.
% while Cutmix and Snapmix based on mixing of patches achieve better results than baseline. 
For a fair comparison, in the auxiliary task ``without rotation" (w/o r), our proposed SePPMix (w/o r) outperforms the baseline by 3.03\% and 2.38\% on 1/5-shot tasks respectively, which significantly surpasses other mix-based methods. Fig.~\ref{fig2} gives some class activation maps of samples obtained using different methods. The proposed SePPMix pays more attention on the object while less on the background, which intuitively demonstrates that our approach yields better results. 

To show that every component in our model has a valid contribution, we use Patchmix to denote that the label is determined proportionally to the area of pixels like Cutmix, while the image is generated the same as ours. The Patchmix performs better than Cutmix, which proves the effectiveness of our patch mix strategy. After combining with our semantic label, SePPMix(w/o r) without rotation improves more than 1\% accuracy compared with Patchmix. Besides, with the auxiliary rotation task, SePPMix(w r) further improves the accuracies to 66.98\% and 83.88\% on 1/5-shot task,  respectively.
\section{CONCLUSION}
In this paper, we aim at better generalizing the model to unseen classes in few-shot learning. To achieve this, we propose a simple but effective mix method called SePPMix, which generates new training data with semantically proportional labels. Additionally, we rotate the generated samples and predict the angles as auxiliary signals. Extensive experiments on the miniImageNet and CUB datasets verify the effectiveness of our method.

%\vfill
%\pagebreak

\vfill\pagebreak

%\section{REFERENCES}
%\label{sec:refs}

  
\bibliographystyle{IEEEbib}
\small

\bibliography{refs}

\end{document}